\documentclass[conference]{IEEEtran}
\IEEEoverridecommandlockouts
\usepackage{cite}
\usepackage{amsmath,amssymb,amsfonts}
\usepackage{algorithmic}
\usepackage{graphicx}
\usepackage{textcomp}
\usepackage{xcolor}
\usepackage{textcomp}
\usepackage{booktabs}
\usepackage{color}
\usepackage{mathptmx}
\usepackage{array}
\usepackage{color}
\usepackage{enumerate}
\usepackage{enumitem}
\usepackage{etoolbox}
\usepackage{multirow}
\usepackage{mathrsfs}
\usepackage{amsmath}
\usepackage{balance}       
\usepackage{graphics}      
\usepackage[T1]{fontenc}   
\usepackage{txfonts}
\usepackage{mathptmx}
\usepackage[pdflang={en-US},pdftex]{hyperref}
\usepackage{color}
\usepackage{booktabs}
\usepackage{textcomp}
\usepackage[T1]{fontenc}
\def\BibTeX{{\rm B\kern-.05em{\sc i\kern-.025em b}\kern-.08em
    T\kern-.1667em\lower.7ex\hbox{E}\kern-.125emX}}
\begin{document}

\title{Multi-modal Representation Learning for Social Post Location Inference}



\author{
\IEEEauthorblockN{RuiTing Dai\IEEEauthorrefmark{2},
Jiayi Luo\IEEEauthorrefmark{2},
Xucheng Luo\thanks{This work was partially supported by Sichuan Science and Technology Program No.2020YFG0053.\newline\indent \IEEEauthorrefmark{1}Corresponding author.}\IEEEauthorrefmark{2}\IEEEauthorrefmark{1},
Lisi Mo\IEEEauthorrefmark{2}, 
Wanlun Ma\IEEEauthorrefmark{3}, and
Fan Zhou\IEEEauthorrefmark{2}}

\IEEEauthorblockA{\IEEEauthorrefmark{2}School of Information and Software Engineering, \\University of Electronic Science and Technology of China, Chengdu 610054, China}
\IEEEauthorblockA{\IEEEauthorrefmark{3}Swinburne University of Technology, Australia}
\IEEEauthorblockA{Email: rtdai@uestc.edu.cn, jiayi.luo59@gmail.com, xucheng@uestc.edu.cn,\\
morris@uestc.edu.cn, wma@swin.deu.au, fan.zhou@uestc.edu.cn} 
}


\maketitle
\begin{abstract}
Inferring geographic locations via social posts is essential for many practical location-based applications such as product marketing, point-of-interest recommendation, and infector tracking for COVID-19. Unlike image-based location retrieval or social-post text embedding-based location inference, the combined effect of multi-modal information (i.e., post images, text, and hashtags) for social post positioning receives less attention. In this work, we collect real datasets of social posts with images, texts, and hashtags from Instagram and propose a novel Multi-modal Representation Learning Framework (MRLF) capable of fusing different modalities of social posts for location inference. MRLF integrates a multi-head attention mechanism to enhance location-salient information extraction while significantly improving location inference compared with single domain-based methods. To overcome the noisy user-generated textual content, we introduce a novel attention-based character-aware module that considers the relative dependencies between characters of social post texts and hashtags for flexible multi-model information fusion. The experimental results show that MRLF can make accurate location predictions and open a new door to understanding the multi-modal data of social posts for online inference tasks.
\end{abstract}


\section{Introduction}
Determining the location of posts is vital for various online social network applications, such as targeted product marketing, user privacy protection~\cite{hussain-2020-privacy}, crisis/disaster detection, and epidemic investigation (e.g., COVID-19). In the last decade, understanding users' behavior (e.g., mobility and trajectory) and identifying their location by online social media have attracted increasing attention from both academicians and the industry~\cite{su-2020-fgcrec}. However, publicly available and accessible social media data with geo-tags is scarce, e.g., less than 1\% of the Tweet posts come with geolocation tags~\cite{fine_pretwe_humacol}. Moreover, inferring social post location differs from the user or home location localization~\cite{zheng-2018-survey}, whose input is all the posts of a particular user -- the former, in contrast, needs to determine the location based on only one user post. 

Therefore, focusing on the specific social post, recent approaches of post location inference are based on the textual content of positions~\cite{ozdikis-2019-locality}, the network of relationships such as friendship, or hybrid relations~\cite{chong-2017-tweet} containing content, network, time while lacking considering images of posts. With the flourishing development of social media platforms and intelligent handset techniques, images have become essential for social posts, from which we can extract useful geographic information. Despite all this, there are no public social network datasets with rich images available for social post location inference. In addition, as social platforms are usually informal communication platforms, frequently but casually posting is encouraged, inevitably producing noisy messages with abbreviated words, misspellings, and emoji characters. It also brings many challenges in research on post-localization. In order to learn meaningful representations from much social information, some studies utilize multiple sources of information for user's location inference~\cite{luo-2020-overview}. However, the previous methods concerning post's location inference focus on each modality and use single-modal features, which may ignore important relations between different modalities.




To address the aforementioned problems, in this paper, we propose a novel multi-modal representation learning framework, which utilizes multi-head attention fusion to capture inter-modal interaction of different content in social posts for location estimation. We begin by collecting real datasets of social posts with images, texts, and hashtags, then model the features representation learning of each modality. In particular, to extract features hidden in the noisy textual content, we introduce a new attention-based character-aware module that employs character-aware embedding and multi-head self-attention for joint learning together with the word-level representation. Since people, objects, and landscapes are usually mixed in social images, which makes location estimation more difficult, we extract geospatial information by matting and inpainting the coverings that may degenerate the location inference performance. Finally, we integrate a multi-head attention mechanism to enhance location-salient information extraction and jointly learn a multi-modal representation for social-post location inference. The main contributions of this work are as follows:
\begin{itemize}[leftmargin=*]

\item We propose a novel multi-modal representation learning framework (MRLF) that considers the interactions between multi-modal features when performing multi-head attention fusing and feature aggregation. It enables our framework to perform cross-modal fusion before obtaining the independent feature representation of each mode. We will demonstrate that MRLF performs better than existing methods that manipulate fusion after obtaining the features.


\item  We introduce an attention-based character-aware module that models the relative dependencies between characters and combines the word-level representation to jointly learn the textual representation of social post texts and hashtags. Compared to the existing social post-location inference approaches, MRLF can considerably improve prediction accuracy while freeing from the computationally intensive steps of building the user relationship map.

\item As the first attempt to study the problem of inferring locations from multi-modal social post representation learning, we release a new dataset with social post texts, hashtags, and images from Instagram. We hope our effort can foster future studies in online social media localization. We conduct extensive experiments to evaluate the performance of our model on real-world datasets. The results show that our method significantly outperforms the baselines.
\end{itemize}


\section{Related Work}
\label{related_work}

According to the type of source media, existing localization solutions can be classified into three categories: text-based, image-based, and multi-modal methods. 

\noindent \textbf{Text-based Methods.}
Textual post content, such as texts and hashtags, has been widely used in location inferences. Early work~\cite{wing-baldridge-2011-simple} typically used various classification models and probabilistic models to infer the geolocation with the location-relevant words or the meaningful indicative words. Due to the lack of location annotations in social posts, deep learning approaches have been applied to learn social post representation by various end-to-end models~\cite{Tagivisor,ozdikis-2019-locality,chen-2020-location}.
Ozdikis et al.~\cite{ozdikis-2019-locality} analyzed the geographical distribution of tweet texts using kernel density estimation and a set of kernel functions for every term. Tagivisor~\cite{Tagivisor} adopted a random forest classifier to exploit the hashtags in the posts and analyze user-posted locations in a city at a grid level. A recent work~\cite{chen-2020-location} proposed a method combining Bi-LSTM neural network with conditional random field to identify geo-entities, showing promising performance in detecting location information using deep neural networks.

\noindent \textbf{Image-based Methods.}
The images of social posts play an essential role in many location inference tasks, such as recognizing points of interest (POIs)~\cite{jinbao-2022-deep} and trip recommendation~\cite{Deeptrip}. However, few studies explore the abundant information behind social images to predict the locations of social posts. An earlier work~\cite{li-2017-geo} collected the geo-tagged images and corresponding geographic information to create indexes with existing scenes in the dataset, such that the location can be predicted through feature detection and comparison. With the development of deep learning techniques in computer vision, researchers have utilized various deep learning models to capture the correlation of different content in social images, such as visual content and text tags~\cite{wagenseller-2019-location}, towards learning latent features for image location prediction.

\noindent \textbf{Multi-modal Inference Methods.}
Recently, a few efforts have been focused on inferring the locations of posts by exploiting the multi-modality of social posts. However, for the user's location~\cite{chong-2019-fine,hamouni-2019-TF-MF,ZhouINF}, the user's context has been sufficiently analyzed to extract direct and indirect location information for user localization. For example, the work~\cite{chong-2019-fine} fused user posts, user profiles, user timelines, and even user timezone to infer user locations. Another work~\cite{hamouni-2019-TF-MF} used communities of users to figure out the physical places. A recent study~\cite{ZhouINF} modeled user relations with hierarchical graph neural networks and used influence functions to explain the geolocation results. However, previous methods do not focus on inferring the geographical location of posts and without model the rich information of user-generated images.

As images become an essential accessory of social posts and may imply location information, the social post localization task requires more effort to explore the social post content. Since the existing fusion methods consider the single-modal feature, what separates our work from them is that we focus on learning relationships among multiple modals effectively. In this work, we initiate the first attempt to learn social post representations using multi-modal information extracted from text, hashtags, and images for more accurate localization.

\section{Method}
\label{method}
\noindent \textbf{Problem Definition.} The problem of multi-modal location inference can be formalized as follows. Assume each post $p = \left (H, T, I\right )$ is composed of 1) a set $H = \left(h_{1}, h_{2}, ... , h_{n} \right)$ of $n$ hashtags $\left(e.g., \#outdoor\right)$, 2) a set $T = \left(t_{1}, t_{2}, ... , t_{q}\right)$ of the text containing $q$ sentences, and 3) a set $I = \left(i_{1}, i_{2}, ... , i_{m}\right)$ of $m$ images. With the notations provided above, the problem can be formalized as: Given a post $p_{i}$ = $\left (H_{i}, T_{i}, I_{i}\right )$, the model needs to learn the multi-modal feature representations and infer the location $l_{i}$.

\begin{figure}
\centering
  \includegraphics[width=0.95\columnwidth]{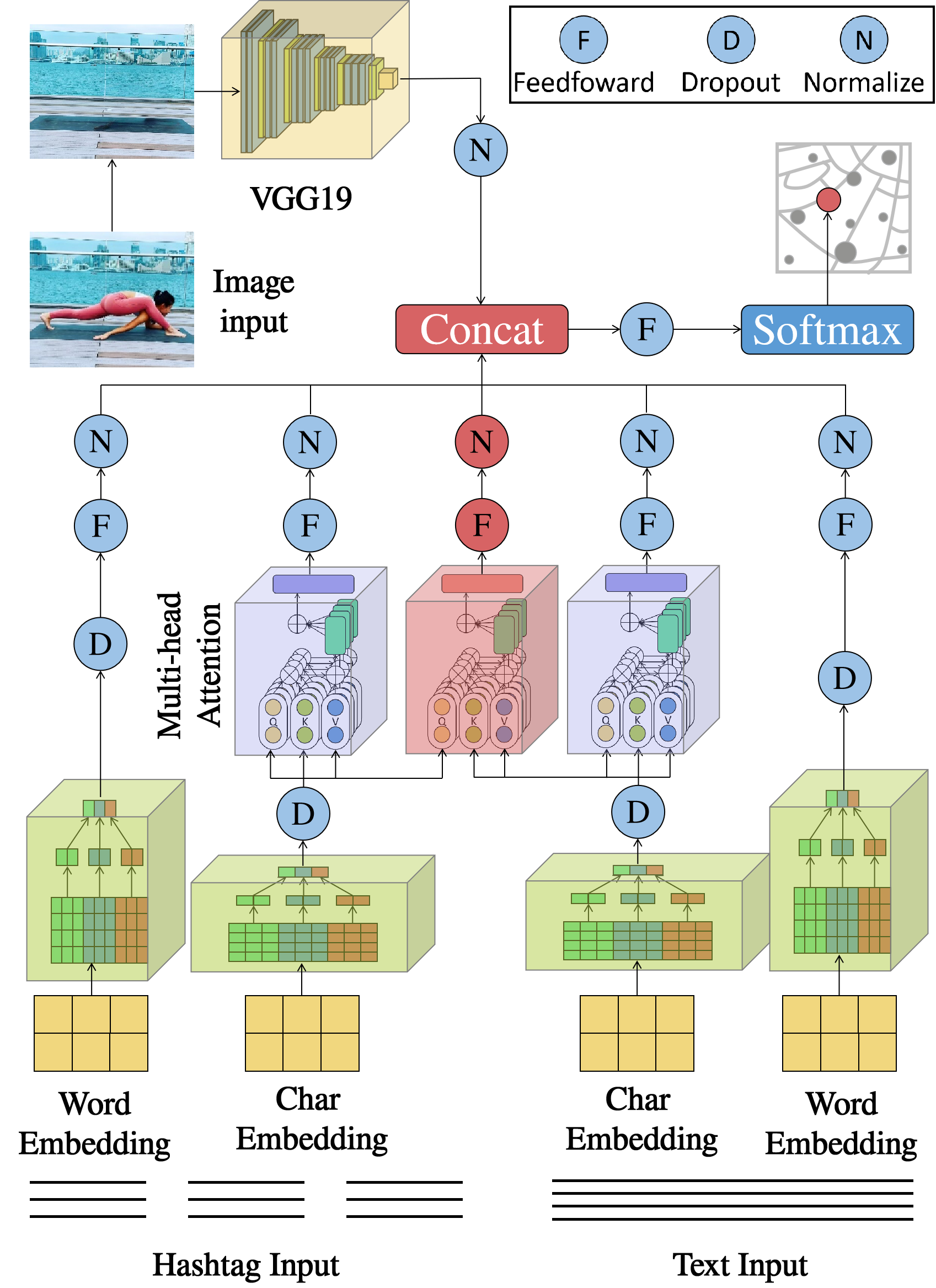}
  \caption{The architecture of MRLF. MRLF takes images, hashtags, and texts related to social posts as input. Images are processed with VGG19. The features of the hashtags and texts are captured with an attention-based network. A convolution layer is then employed to fuse cross-modal features and obtain the multi-modal representations of social posts, as shown in the red part.}~\label{fig:model_architecture}
\end{figure}

\noindent \textbf{Overview of MRLF.} 
As illustrated in Fig.~\ref{fig:model_architecture}, MRLF aggregates the feature representations of images, hashtags, and texts so as to infer the post's locations. Images are processed with a VGG19-based model. Meanwhile, an attention-based character-aware network is proposed to extract the implicit representations of both hashtags and texts, while a convolution-based network is used to capture the explicit representations at the word level. In contrast to prior approaches that use the attention mechanism to capture cross-modal features in the final stage, we advance the fusion stage to the convolutional layer. Finally, we use multi-modal feature representations to infer a post's location.

\subsection{Hashtag and Text Feature Representation Learning}
Users often inadvertently reveal their posting location through text or hashtags, both of which may conceal inferable location information. Given a cluster of characters, we use one-hot to build a vocabulary for all characters or words encoded as vectors with a fixed dimension $l$. To capture the implicit meanings of the low-frequency and informal words of the post, we add a character-aware embedding combined with the word embedding, which is also used to infer the explicit meaning of hashtags and texts. Here, we use a convolution-based model to extract the feature of word embedding. At the same time, we adopt a multi-head self-attention-based module to model the character-level features.

\subsubsection{Character-level feature representation}
Attention-based models usually perform well in extracting correlations by quantifying the interdependence among the inputs. However, it will cause intensive computations in our model. To capture the significant property while reducing the sequence length, we address the computation issue by applying a 1-D convolution neural network and, subsequently, a maximum pooling layer over a small window to extract the information. Given the character matrix $C$, each element of the result matrix $H^{max}$ is calculated as:
\begin{align}
    h_{ij}^{conv} &= C\left(i-k:i+k\right)W_{j}^{conv} + b_{j}^{conv}, \\
    h_{in}^{max} &= \max\left(H^{conv}\left(l-k:l+k, j\right)\right),
\end{align}
where $k$ is half of the kernel size, $i$ is the index of the character sequence ranging from $1$ to the character length $m$, $j$ is the index of the filter ranging from $1$ to the number of filters, $b_{j}^{conv}$ is the bias, $h_{ij}^{conv}$ is the each element of matrix $H^{conv}$, $l = i\left(2k-1\right)$, $n$ rangs from $1$ to $m/\left(2k-1\right)$, and $h_{in}^{max}$ is the each element of matrix $H^{max}$. After obtaining $H^{max}$, we utilize a linear function to transform the matrix $H^{max}$ into query matrix $Q$, key matrix $K$ and value $V$, and split each matrix to $h$ parallel heads:
\begin{align}
    &O_{i} = softmax\left(\frac{\left(Q_{i}W_{i}^{Q}\right)\left(K_{i}W_{i}^{K}\right)}{\sqrt{d/h}}\right)\left(V_{i}W_{i}^{V}\right),\\
    &O = concat\left[O_{1},\cdot\cdot\cdot,O_{i},\cdot\cdot\cdot,O_{h}\right],
\end{align}
where $i\in\left(1,h\right)$, $O$ is the attention score matrix, and $d$ is the dimension of the matrix. Finally, the character representation $F^{c}$ can be computed with $W_{1}^{c}$, $W_{2}^{c}$, where $b_{1}^{c}$ and $b_{2}^{c}$ are all learnable parameters:
\begin{align}
    F^{c} = \max\left(OW_{1}^{c} + b_{1}^{c}, 0\right)W_{2}^{c} + b_{2}^{c},
\end{align}

\subsubsection{Word-level feature representation}
To obtain explicit feature representations, we transform each text and hashtag sequence into a vector sequence at the word level and generate the matrix representations of text and hashtag sequences. It is worth noting that we remove low-frequency hashtags since the meaning of these hashtags would significantly mislead the geolocation information extraction, which we will discuss in detail in Sec.~\ref{low-frequency hashtags}. Similar to the regular operation in computer vision, we also employ convolution and max-pooling to learn the word-level feature representations of hashtag sentences and texts.

\subsection{Image Feature Representation Learning}
Images consist of noise or location-irrelevant information (e.g., selfies and pet photos), which contributes less and even negatively to location inference. To alleviate the adverse impact of image noise, we recognize portraits~\cite{qin-2020-u2} in images and perform image fill inpainting~\cite{suvorov-2022-resolution}. In Sec.~\ref{noisy images}, we provide the detailed experimental results of deleting image noise and its impact on the model predictions. When implementing MRLF, we use a pre-trained VGG19 model~\cite{simonyan-2015-very} to generate image feature representation, which adopts a combination of small filter convolution layers. In particular, we freeze all pre-training parameters in the convolutional layer and apply a fully connected layer following a global average pooling layer to adjust the size of the image feature representation vector.

\subsection{Multi-modal Representation Learning}
Humans usually infer location by the same words in text and hashtags. To mimic this nature of humans, we advance the fusion of text and hashtags character features to the convolutional layer. Based on the output matrix $H_{text}^{max}$ and $H_{tag}^{max}$ of texts and hashtags in the convolution layer, we obtain the eventual multi-modal feature fusion as
\begin{align}
    &O_{cro} = softmax\left(\frac{\left(H_{tag}^{max}W_{tag}^{Q}\right)\left(H_{text}^{max}W_{text}^{K}\right)}{\sqrt{d_{text}}}\right)\left(H_{text}^{max}W_{text}^{V}\right), \\
    &A_{cro} = \max\left(O_{cro}W_{cro}^{1}+b_{cro}^{1},0\right)W_{cro}^{2}+b_{cro}^{2},\\
    &F_{cro} = LayerNorm\left(A_{cro}\right),
\end{align}
where $F_{cro}$ is the fused representation of text and hashtags. We explain the reasons for not fusing image features in Sec.~\ref{multi-modal fusion}. Lastly, we obtain the multi-modal geographic information feature representation by concatenating the representations derived from the different modals mentioned above:
\begin{align}
    F_{p} = concat\left[F_{text}^{c},F_{text}^{w},F_{tag}^{c},F_{tag}^{w},F_{img},F_{cro}\right],
\end{align}
where $F_{text}^{c}$ and $F_{tag}^{c}$ are the character-level representations of texts and hashtags, $F_{text}^{w}$ and $F_{tag}^{w}$ are the word-level representations of texts and hashtags, $F_{img}$ are the feature representations of images.

\subsection{Location Inference}
By applying transformation matrix $W_{p}$ to shape the output dimension, we could get the vector for prediction:
\begin{align}
    V = F_{post}W_{p},
\end{align}
where $V\in\mathbb{R}^{m_{l}}$ and $m_{l}$ is the location size. We then apply the softmax function to calculate the probability for each location:
\begin{align}
    p_{i} = \frac{e_{i}}{ {\textstyle \sum_{j=1}^{m_{l}}}e_{j}},
\end{align}
where $e_{i}$ is the $i-$th element of $V$ and $i$ is ranging from $1$ to $m_{l}$. The prediction of location $l'$ is the label with the highest probability:
\begin{align}
    l' = \arg\max\left(p_{1},\cdot\cdot\cdot,p_{i},\cdot\cdot\cdot,p_{m_{l}}\right).
\end{align}

\noindent \textbf{Loss Function.}
During training, we use cross entropy as the loss function:
\begin{align}
    Loss = \frac{1}{m_{l}}\sum_{i}^{m_{l}}l_{i}\log\left(l'_{i}\right),
\end{align}
where $l'_{i}$ is denotes the probability of $i-$th location and $l_{i}$ are binary indicators.


\section{Experiments}
\label{experiments}

\subsection{Experimental Settings}
\noindent \textbf{Data Collection.}
For multi-modal localization method evaluation, it is necessary to use social post data with rich images, text, and hashtags information. Therefore, we curate three multi-modal datasets from Instagram, the most popular online social networking platform for image sharing. Specifically, we select three popular cities on Instagram: New York, Melbourne, and Hong Kong. Detailed statistics of the three datasets are summarized in Table~\ref{table: dataset statistics}.

\begin{table}[ht]\small
    \caption{Dataset Statistics.}
    \label{table: dataset statistics}
    \renewcommand\arraystretch{1.2}
    \centering
    \setlength{\tabcolsep}{3.7mm}{\begin{tabular}{lrrr}
\toprule[1.2px]
\textbf{Category} & \textbf{New York} & \textbf{Melbourne} & \textbf{Hong Kong} \\ \hline
\#Posts             & 137,676           & 43,656               & 43,393           \\
\#Locations         & 40                & 20                   & 27               \\
\#Images            & 16,357            & 12,956               & 38,414           \\
\#Hashtags          & 137,676           & 33,296               & 33,525           \\
\#Text              & 116,961           & 30,529               & 37,235           \\
\bottomrule[1.2px]
\end{tabular}}
\end{table}

\noindent \textbf{Data Preprocessing.} 
We find the locations of popular check-ins in each city using top-pick search results through Foursquare. Then we collect the Instagram data using the Instaloader library\footnote{https://instaloader.github.io}. As social media data is always irregular and cluttered, several filtering strategies are applied to the collected data. First, we remove the hashtags that appear less than 50 times in each city. In image segmentation and recognition, we set a parameter $\eta$ to filter out pictures with excessive noise information, such as selfies. In the experiments, we set $\eta = 0.5$ to delete useless photos. In each city, we only keep geographic locations with at least 100 check-in posts, which excludes the POIs that people rarely visit or mention. Specifically, we selected English posts that contain text, hashtags, and images in the dataset for MRLF training.

 

\noindent \textbf{Baselines}. \label{section:Baselines}We compare MRLF with two categories of baselines. The first category consists of text-based or hashtag-based methods, including:
 \begin{itemize}[leftmargin=*]
    \item \textbf{Tagvisor}~\cite{Tagivisor} utilizes the random forest model to mine the potential connections between hashtags and user locations for predictions. 
    \item \textbf{Deepgeo}~\cite{lau-etal-2017-end} uses a Bi-LSTM with the attention mechanism to learn the relationships between different indicative word vectors and discriminate the locations of the posts.
    \item \textbf{LocTwi}~\cite{huang-2019-location} treats subwords as a feature and further improves the representation learning of informal language on social networks. 
\end{itemize}

The second category consists of image-based methods. Location inference for images can be treated as a classification task, but few works have been done in the literature. We choose several representative image classification models for comparison, including:
 
\begin{itemize}[leftmargin=*]
\item \textbf{ResNet50}~\cite{he-2016-deep} is a variant of ResNet, which has 48 Convolution layers along with 1 MaxPool and 1 Average Pool layer. 
\item \textbf{ResNet101}~\cite{he-2016-deep} is a deeper modal based on ResNet50.
\item \textbf{DenseNet121}~\cite{huang-2017-densely} connects each convolution layer to every other layer in a feed-forward fashion.
\end{itemize}

\noindent \textbf{Evaluation metrics.}
We use two metrics to evaluate all methods following previous related works~\cite{chong-2017-tweet,ZhouINF}: (1) \textit{Mean} is the averaged errors between the predicted cluster centers and the ground-truth geolocations, and (2) \textit{Acc} measures the accuracy of the classification results.

\noindent \textbf{Parameter setting.}
We set the maximum number of hashtags and text characters to $100$. Besides, the embedding dimensions of text and hashtags are set to $100$. We apply convolutions on both character-aware and word representation learning. The filter sizes are $\left(3,4,5,6\right)$ and $\left(1,2,3,4\right)$, respectively, and the filter number is set as $100$. We set the number of heads in the multi-head self-attention mechanism to $8$. When proceeding to the feedforward layer, we also add a dropout to stabilize the training of our model, whose parameter is set to $0.5$. We adopt the Rmsprop optimizer to train the model with a learning rate of $0.001$, and the weight decays by 0.8 per time with patience after $10$ epochs. The model converges after we train $100$ epochs. All the neural network models are implemented using Tensorflow. All experiments are conducted on a machine with an Intel(R) Xeon(R) Silver 4110 CPU and two GeForce RTX 3090 GPUs.


\begin{table}[ht]\small
    \caption{Performance comparisons on three datasets.}
    \label{table: comparison experiment}
    \renewcommand\arraystretch{1.3}
    \centering
    \setlength{\tabcolsep}{1.5mm}{\begin{tabular}{lllllll}
\toprule[1.2px]
\multirow{2}{*}{Model} & \multicolumn{2}{l}{New York}      & \multicolumn{2}{l}{Melbourne}     & \multicolumn{2}{l}{Hong Kong}     \\ \cline{2-7} 
                       & Acc             & Mean             & Acc             & Mean             & Acc             & Mean             \\ \hline
Tagivisor~\cite{Tagivisor}              & 0.681          & 2.385          & 0.652          & 0.724          & 0.523          & 1.745          \\
Deepgeo~\cite{lau-etal-2017-end}                & 0.738          & 1.804          & 0.742          & 1.243          & 0.654          & 1.522          \\
LocTwi~\cite{huang-2019-location}                & 0.724          & 1.921          & 0.734          & 1.220          & 0.644          & 1.582          \\
ResNet50~\cite{he-2016-deep}               & 0.669          & 2.785          & 0.694          & 1.257          & 0.811          & 0.767          \\
ResNet101~\cite{he-2016-deep}              & 0.644          & 3.608          & 0.718          & 1.368          & 0.809          & 0.730          \\
DenseNet121~\cite{huang-2017-densely}         & 0.553          & 3.927          & 0.604          & 1.608          & 0.538          & 1.892          \\ \hline
\textbf{MRLF}          & \textbf{0.828} & \textbf{1.510} & \textbf{0.860} & \textbf{0.512} & \textbf{0.847} & \textbf{0.715} \\ \bottomrule[1.2px]
\end{tabular}}
\end{table}
 
\subsection{Performance Comparison}
The overall performance comparison of all methods across three datasets is presented in Table~\ref{table: comparison experiment}, from which we have the following observations. First, MRLF consistently outperforms baselines on all metrics, e.g., the performance gains of MRLF over the best baseline method in terms of Accuracy and Mean are $9.0\%$ and $0.294km$ in New York, $11.8\%$ and $0.212km$ in Melbourne, and $3.6\%$ and $0.015km$ in Hong Kong, respectively. It further demonstrates that the fusion of multi-modal features helps improve social content location inference performance. Next, we can observe that different components play distinct roles in different datasets, as the POIs of each dataset have various visual distinctions, and people in other regions have different habits of using social networks. Third, models achieving higher Accuracy performance do not mean smaller Mean results, e.g., Tagivisor and Deepgeo in Melbourne, ResNet50 and ResNet101 in Hong Kong, and ResNet101 and VGG16 in New York. For the Mean metric, incorrectly predicting two POIs far from each other may counteract the contribution made by multiple correct predictions.
 
\subsection{Ablation Study}
\begin{figure}[ht]
 \centering
  \includegraphics[width=1\columnwidth]{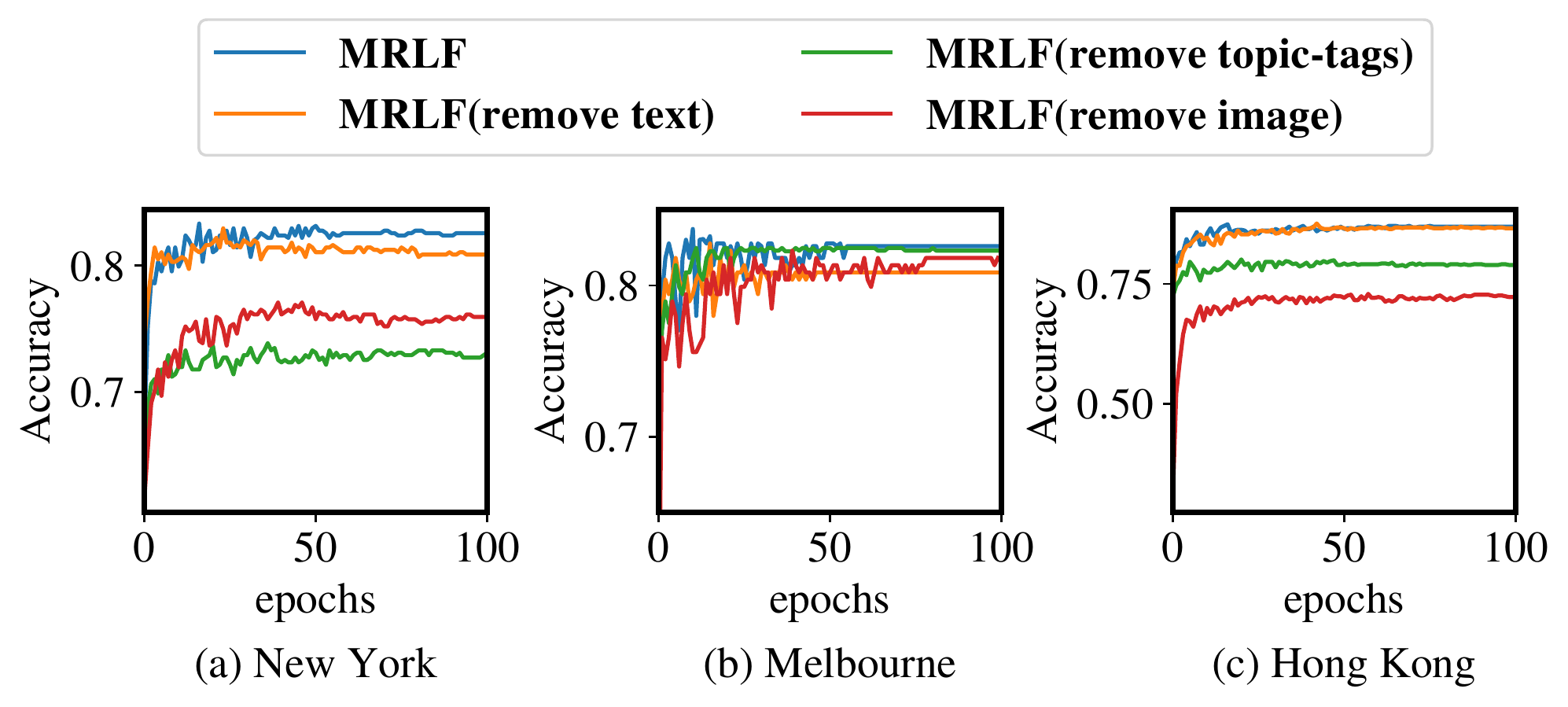}
  \caption{The ablation results investigating the influence of different components. We only report accuracy results due to space limits, but the same trend is also hold for Mean evaluations.}~\label{fig:ablation}
\end{figure}

We conduct an ablation study to quantify the effect of each component in our MRLF. As shown in Fig.~\ref{fig:ablation}, we study all three critical features in our model: the \textit{text}, \textit{hashtags}, and the \textit{image} representations. Text and hashtags are less important in Melbourne and Hong Kong datasets, but MRLF still outperforms other methods. The result also suggests that the performance of MRLF depends on all the components, i.e., the three modalities contribute to the final prediction, which proves the effect of multi-modal information fusion for geographic prediction. More importantly, the image information contributes more than the other two -- an important finding that may inspire future work in online content localization.

\subsection{Multi-modal Fusion Study}
\label{multi-modal fusion}
\begin{figure}[ht]
 \centering
  \includegraphics[width=1\columnwidth]{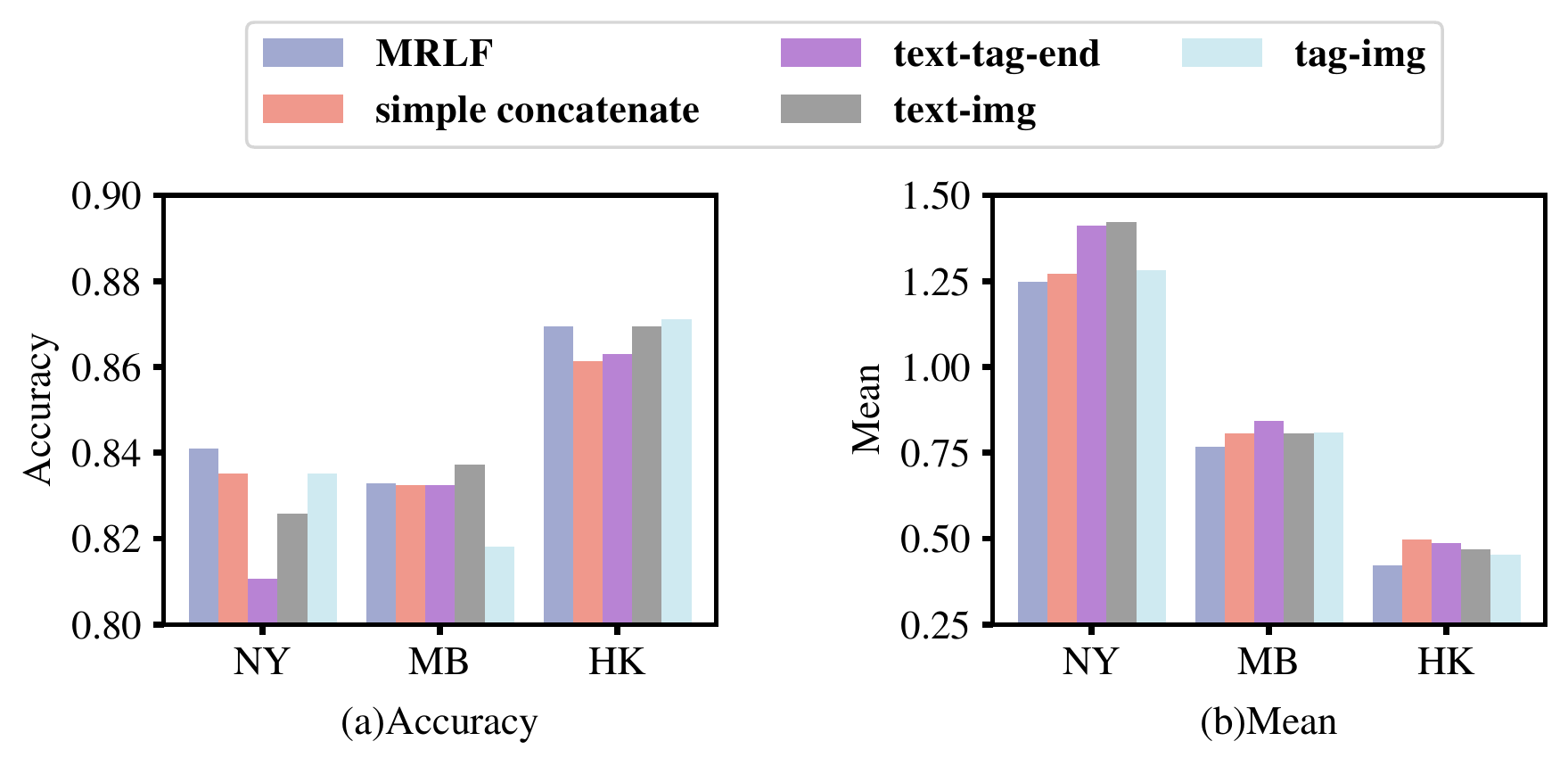}
  \caption{Comparative experimental results of fusing different components.}~\label{fig:fusion}
\end{figure}

 Since many existing methods are based on attention mechanisms for cross-modal feature fusion, we now investigate the influence of different fusion methods. As shown in Fig.~\ref{fig:fusion}, we compare the performance of hashtags and text fusion before and after getting each feature, which corresponds to the \textbf{MRLF} and \textbf{text-tag-end} in Fig.~\ref{fig:fusion}, respectively. MRLF achieves better performance due to the cross-modal fusion before getting isolated features, which encourages the convolution layer to pay more attention to the commonality of the two components. This result is like humans usually infer location by the same word in both hashtags and text. An important finding is that the fusion of different components and concatenation simply has similar results. Even the fusion of text and images performs slightly higher than MRLF on the Melbourne dataset. However, MRLF outperforms all fusion methods in all datasets in terms of Mean evaluation because only a small percentage of images in social networks have road signs, which may reduce the performance of fusing the image with other components.

\subsection{Influence of Noisy Images}
\label{noisy images}
Generally, social media images consist of much personal information that is geographic-free. Therefore, user-generated contents usually include a variety of close-ups of individual items. For example, selfies are a kind of noise for capturing geographical information. In Table~\ref{table: remove_nosie_image}, we compared the performance before and after removing noisy images with a portrait ratio greater than 0.5. In particular, we use all the posts in the dataset and only adopt the image feature representation learning part in MRLF for training. 
As the results demonstrated, after removing noisy images, the Mean performance is significantly improved on all datasets, and higher accuracy is also achieved on the Melbourne dataset. This result verifies our motivation for removing geographic-free noise in multi-modal fusion and post-localization.

\begin{table}[ht]\small
    \caption{Comparison of performance before and after noise image removal.}~\label{table: remove_nosie_image}
    \renewcommand\arraystretch{1.3}
    \centering
    \setlength{\tabcolsep}{1.6mm}{\begin{tabular}{lllllll}
\toprule[1.2px]
\multirow{2}{*}{Method} & \multicolumn{2}{l}{New York} & \multicolumn{2}{l}{Melbourne} & \multicolumn{2}{l}{Hong Kong} \\ \cline{2-7} 
                        & Acc           & Mean          & Acc           & Mean           & Acc           & Mean           \\ \hline
Image (all)               & 0.579         & 3.755        & 0.676         & 1.399         & 0.600         & 1.708         \\
Image (remove)      & 0.579         & 3.695        & 0.697        & 1.261         & 0.600         & 1.669          \\ \bottomrule[1.2px]
\end{tabular}}
\end{table}

\subsection{Influence of Low-frequency Hashtags}
\label{low-frequency hashtags}
\begin{figure}[h]
\centering
  \includegraphics[width=1.0\columnwidth]{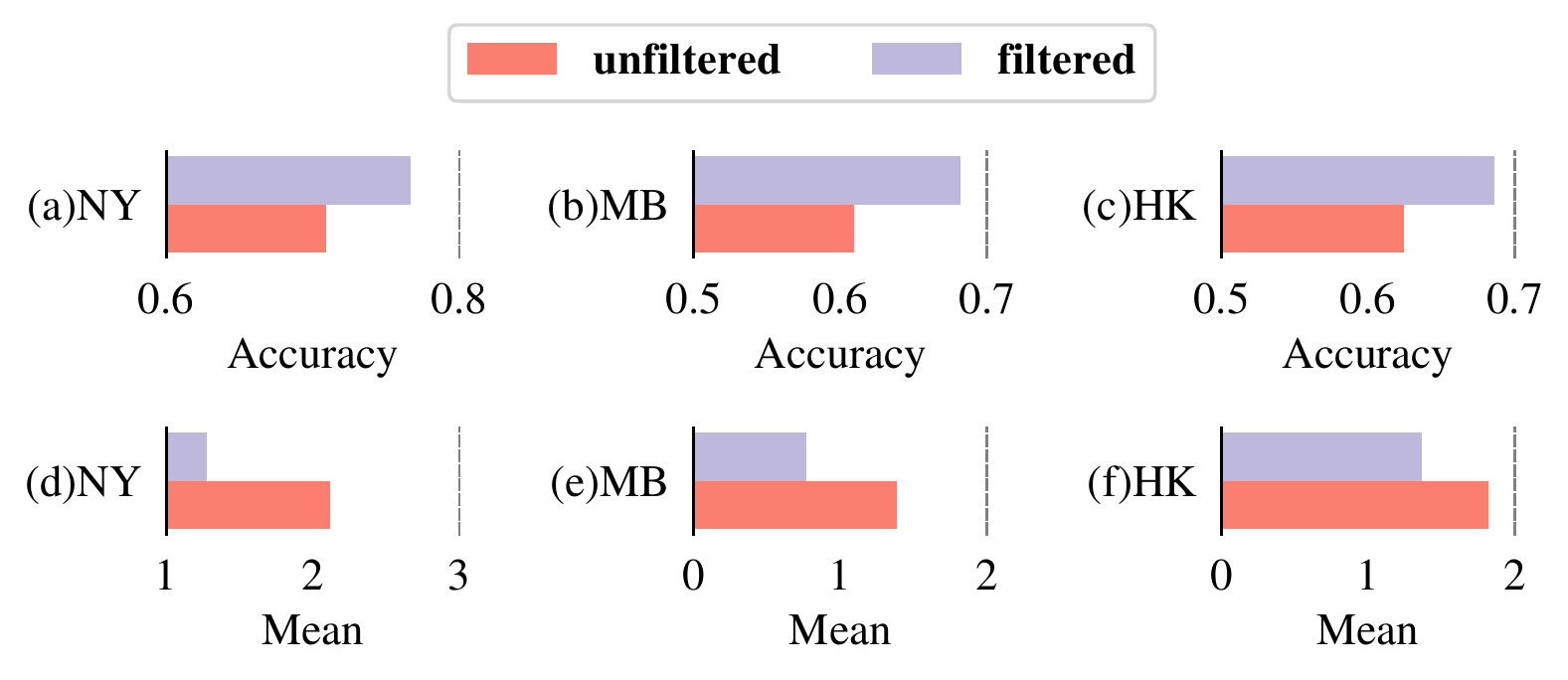}
  \caption{The impact of low-frequency hashtags on prediction performance.}~\label{fig:filter_tag}
\end{figure}

As shown in Fig.~\ref{fig:filter_tag}, we quantify the influence of filtering the low-frequency hashtags. After filtering, the model performs better under the two metrics. This phenomenon suggests that word vectors that are insufficiently trained may mislead the extraction of valid knowledge and, therefore, deteriorate the prediction performance. Removing these data that do not have access to sufficient training can improve the performance of the model prediction.

\section{Conclusions}
\label{conclusion}
This paper exploits just one social post for the geolocation prediction task. To the best of our knowledge, this is the first post-localization study that leverages complete post information, including texts, hashtags, and images. The proposed model MRLF integrated a multi-head attention mechanism to enhance location-salient knowledge for multi-modal representation fusion. In addition, we presented an attention-based character-aware module that not only perceives the relative dependency of character-level features but also fuses the character-level features before obtaining the text and hashtag features to learn the textual representation better. Further, we improved the model's performance by removing the effect of noise from both hashtags and images. Experimental results on real-world datasets curated from Instagram verified the effectiveness of our solution. It is interesting to use neural networks for multi-modal fusion in our future work.

\vspace{12pt}
\bibliographystyle{IEEEtran}
\bibliography{main} 

\end{document}